\documentclass{article} 
\usepackage{iclr2027_conference,times}

\usepackage{amsmath,amsfonts,bm}

\def\eqref#1{equation~\ref{#1}}

\def\1{\bm{1}}

\DeclareMathAlphabet{\mathsfit}{\encodingdefault}{\sfdefault}{m}{sl}
\SetMathAlphabet{\mathsfit}{bold}{\encodingdefault}{\sfdefault}{bx}{n}

\def\gI{{\mathcal{I}}}

\newcommand{\Ls}{\mathcal{L}}
\newcommand{\R}{\mathbb{R}}

\usepackage[table]{xcolor}

\usepackage{booktabs}
\usepackage{multirow}
\usepackage{amssymb}
\usepackage{pifont}
\usepackage{enumitem}
\definecolor{markgreen}{RGB}{77,169,90}
\definecolor{markred}{RGB}{235,70,47}
\newcommand{\cmark}{\textcolor{markgreen}{\ding{51}}}
\newcommand{\xmark}{\textcolor{markred}{\ding{55}}}
\newcommand{\hd}[1]{\textbf{#1}}          
\newcommand{\vhd}[1]{\multirow{2}{*}{\hd{#1}}} 
\renewcommand{\arraystretch}{1.1}
\definecolor{oursrow}{HTML}{FDE9E1}
\definecolor{baserow}{HTML}{EAF6E5}
\newcommand{\venue}[1]{\,{\tiny\textit{(#1)}}}
\newcommand{\up}{$\scriptstyle\uparrow$}
\newcommand{\down}{$\scriptstyle\downarrow$}
\newcommand{\grouphead}[2]{\multicolumn{#1}{l}{\textit{#2}}\\\cmidrule{1-#1}}
\newcommand{\grouprule}[1]{\cmidrule{1-#1}}
\newcommand{\tablestyle}{\par\vspace{4pt}\footnotesize\setlength{\aboverulesep}{0pt}%
  \setlength{\belowrulesep}{0pt}\renewcommand{\arraystretch}{1.25}}
\usepackage{graphicx}
\usepackage{tikz}
\usepackage{wrapfig}
\usepackage{placeins}   
\newlength{\panelsep}
\newcommand{\panelfont}{\small\bfseries}

\NewDocumentCommand{\fpanel}{O{0.3\textwidth} O{0pt} m m}{%
  \begin{tikzpicture}[baseline=(img.north)]
    \node[inner sep=0pt, outer sep=0pt, anchor=north] (img) at (0,0)
      {\includegraphics[width=#1]{figure/#3}};
    \node[inner sep=0pt, outer sep=0pt, anchor=north, font=\panelfont,
          yshift=-\panelsep, xshift=#2] at (img.south) {(#4)};
  \end{tikzpicture}%
}

\usepackage{hyperref}
\usepackage{url}

\newsavebox{\pairbox}   


\title{What Visual Generators Need from Teachers: Rethinking Representation Alignment}
\author{%
\begin{minipage}[t]{\dimexpr\textwidth-2\tabcolsep\relax}
\raggedright
\textbf{Yongcong Wang}$^{1}$ \quad
\textbf{Hingchin Chen}$^{2}$ \quad
\textbf{Mingyu Fan}$^{3}$ \quad
\textbf{Shuo Jiang}$^{4}$ \quad
\textbf{Teer Zhang}$^{5}$ \\[3pt]
\textbf{Yucong Sun}$^{5,6}$ \quad
\textbf{Zijia Wang}$^{7,8,9}$ \quad
\textbf{Yiming Lu}$^{10}$ \quad
\textbf{Chengchao Shen}$^{1}$\thanks{Corresponding author.} \\[6pt]
\normalfont\small
$^{1}$Central South University \quad
$^{2}$The Hong Kong University of Science and Technology \\
$^{3}$Tsinghua University \quad
$^{4}$The Chinese University of Hong Kong, Shenzhen \\
$^{5}$SenseTime Research \quad
$^{6}$Shandong University \quad
$^{7}$Imperial College London \\
$^{8}$University of Oxford \quad
$^{9}$Dell Technologies \quad
$^{10}$University of International Relations \\[4pt]
\texttt{ycwang1031@gmail.com, scc.cs@csu.edu.cn}
\end{minipage}
}

\iclrpreprintcopy
\begin{document}

\maketitle

\begin{abstract}
Representation alignment speeds up diffusion transformer training by pulling an intermediate block
of the model (student) toward features of a frozen pretrained encoder (teacher). Which
teacher layer to align, and for how long, is still set by convention, and each alternative costs a
training run. We find that alignment helps where the student cannot linearly recover the teacher's
features, not where it already resembles them. Since a deep teacher layer is largely predictable
from the one below, we isolate what each layer adds, its \emph{increment}, and measure how much of
it an unaligned student recovers. The student fills the teacher's hierarchy from the bottom up and
stalls near the top, which we call \emph{hierarchy filling}: even after 400K steps it recovers almost
none of the deepest. The \emph{recoverability gap} is the unrecovered share of an
increment, read from one unaligned checkpoint. In short runs that each align one teacher layer at
one block, the gap nearly reproduces their ranking by FID improvement, and CKA, a measure of feature similarity, largely reverses it. \textbf{R}epresentation \textbf{A}lignment and \textbf{R}ecoverability
\textbf{E}stimation (RARE) picks the teacher layer with the largest gap before training. During
training, it tracks each token's remaining distance to that layer, the online counterpart of the
gap, weights tokens by it, and phases out the loss once the average distance stops falling. With
SiT-B/2 on ImageNet $256\times256$, RARE reaches an FID of $18.02$ without guidance and $4.46$ with
it, ahead of seven alignment baselines including REPA, iREPA and HASTE. It also trains in 14\% fewer
GPU-hours than iREPA. Its FID stays below iREPA's across model scales, teachers, datasets and
backbones.
\end{abstract}

\section{Introduction}

\begin{figure}[t]
\centering
\includegraphics[width=0.9\textwidth]{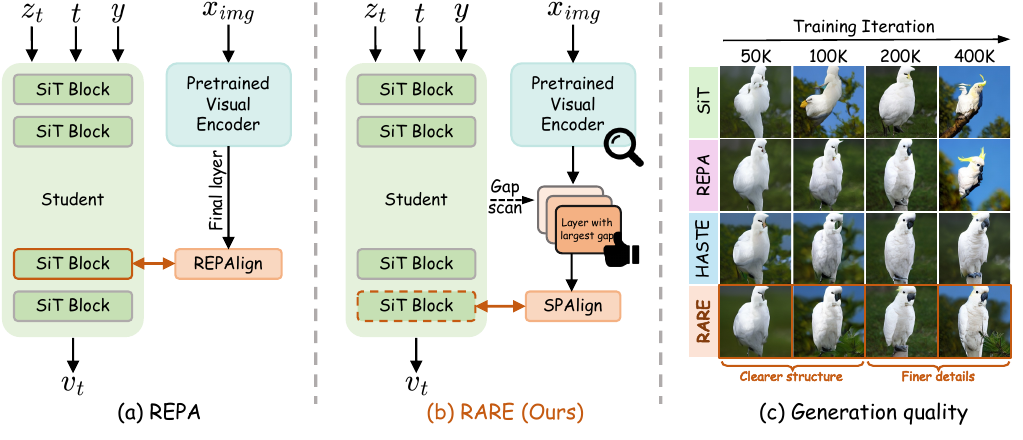}
\vspace{-13pt}
\caption{\textbf{Alignment placed by convention and by measurement.} \textbf{(a)}~REPA alignment
(REPAlign) pulls a fixed student block toward the teacher's deepest layer for the whole run.
\textbf{(b)}~RARE first reads from a frozen checkpoint how much of each teacher layer the student
still lacks, then applies sparse alignment (SPAlign): it aligns only the layer with the largest
gap, puts more weight on tokens that are still far from the target, and switches the loss off once
this distance stops falling, so alignment covers one layer and only the early part of training.
\textbf{(c)}~One class and noise draw at four training steps, guidance scale 4.0.}
\label{fig:teaser}
\end{figure}

Aligning an intermediate block of a diffusion transformer to a frozen visual encoder, as REPA
\citep{repa} does, has become a standard way to speed up generative training. The recipe REPA
set, and most follow-ups inherit, fixes three choices (Figure~\ref{fig:teaser}a): the teacher's
deepest layer as the target, a mid-depth student block, and an alignment loss kept on for the
whole run. Later work has revisited these choices, and the encoder and the loss besides, but each
search trains one run per candidate, and its answer holds for one model and one teacher. What the
benefit of alignment depends on is still open. It is not the representation alone: \citet{raev2} find that a DINOv2 feature already serving as the latent still improves
training when aligned again at an intermediate block. Whether a signal helps depends on what the
student has and has not built where the signal arrives.

The criteria proposed so far do not measure this. They score a property of the teacher's
features, such as linear-probe accuracy in REPA or spatial structure in iREPA \citep{irepa}, how
much the loss depends on a student block \citep{agrepa}, the loss of a short run \citep{condres},
the agreement between alignment and denoising gradients \citep{haste}, or how closely a block
already resembles the teacher. None asks what the student lacks, although a signal can only add what
the student has not built. We test them on fourteen short training runs, each aligning one teacher
layer at one student block (a \emph{placement}), under identical settings
(Section~\ref{sec:criteria}). None of these criteria orders the runs by their FID improvement, and
similarity measures such as CKA \citep{cka} order them in reverse: the more a block already
resembles a teacher layer, the less aligning it to that layer helps.

To see why similarity points the wrong way, we isolate what each teacher layer adds beyond the
layer below it, its increment, and ask how much of it a frozen student can linearly recover at each
block and noise level. The generative objective fills the hierarchy from the bottom up and stalls
near the top (Figure~\ref{fig:filling}): shallow increments are recovered early, while the deepest
stay at the noise level of the measurement through 400K steps, across model sizes, teachers and
image domains. We
call this regularity \emph{hierarchy filling}. It explains the reversal: a block resembles the
teacher where the objective has already filled the hierarchy, and the stalled top, where REPA's
default target lies, is where an external signal has something to add.

Measuring the stall gives a placement criterion. The \emph{recoverability gap} is the share of a
teacher increment that the student cannot linearly recover at a given block and noise level,
relative to the best readout the measurement achieves anywhere. Read from one checkpoint of the
unaligned run, it orders the fourteen runs by their FID improvement (Spearman $\rho=0.92$), whereas
the slope of the denoising loss toward the same target does not: benefit follows how far the
student is from a layer, not how steep the first step is. RARE turns the gap into a
training recipe (Figure~\ref{fig:teaser}b): it aligns the teacher layer with the largest gap, weights
each token by how far its projected feature still is from the target, and switches the loss off
once this distance stops falling. Nothing is added at inference.

With SiT-B/2 on ImageNet $256\times256$, RARE reaches an FID of $18.02$ without guidance and
$4.46$ with it, ahead of all seven alignment baselines, in 14\% fewer GPU-hours than iREPA, and it
improves on iREPA across model scales, teachers, image domains and backbones. In summary:
\begin{itemize}[leftmargin=*,itemsep=1pt,topsep=2pt]
\item We show that the generative objective fills the teacher's hierarchy from the bottom up: a
student trained without alignment linearly recovers what shallow teacher layers add, but almost
none of what the deepest layers add (Section~\ref{sec:filling}).
\item We propose the recoverability gap, read from one unaligned checkpoint, and show that it orders
placements by alignment benefit, while similarity-based criteria order them in reverse
(Section~\ref{sec:gap}).
\item We introduce RARE, which aligns the layer with the largest gap, weights tokens by their
distance to the target and switches the loss off once that distance plateaus; it outperforms REPA,
iREPA and HASTE at lower training cost (Sections~\ref{sec:rare} and~\ref{sec:experiments}).
\end{itemize}

\section{Related Work}

\textbf{Representations in generative models.} Pretrained visual representations now enter the
generative pipeline at every stage. At the tokenizer, VA-VAE \citep{vavae} regularizes the latent
toward a vision foundation model, RAE \citep{rae} replaces the autoencoder with a frozen encoder,
and REPA-E \citep{repae} trains tokenizer and denoiser end to end through an alignment loss. As
intermediate supervision, REPA aligns noisy hidden states to clean teacher features, and later
work changes the target to relational structure \citep{srepa}, to the model's own deeper layers
\citep{sra}, or to a representation entangled with the latent \citep{reg}. At the output,
representation-space distances serve as the training loss \citep{fdloss}, and perception tasks are
posed as image generation \citep{visionbanana}. These designs share a premise: the denoising
objective learns discriminative features on its own but lags self-supervised encoders
\citep{ddae,ldae,rcg}. A representation present somewhere in the system is not necessarily usable
at a given computation stage, as RAEv2 shows for diffusion and studies of language models report
\citep{availableused,diagnosticnotcausal}. We ask which part of a teacher's hierarchy the student
lacks at a given block and noise level.

\textbf{Choosing the target, block and schedule.} Teacher-side studies compare encoders. REPA
favors linear-probe accuracy, and iREPA finds across 27 encoders that spatial structure predicts
generation quality far better than classification accuracy. RAEv2 reports that the relevant
property depends on whether the representation serves as the latent or as the target, and weighted
diversity \citep{diversedit2} and spectral energy \citep{spectrummatching} are further candidates.
All of them align the deepest layer of the chosen encoder. Student-side studies choose the block
and the schedule. Layer sweeps in REPA, U-REPA \citep{urepa} and AHPA \citep{ahpa} settle on
mid-depth blocks, conditioning residuals \citep{condres} select the depth by a short run's loss, and
AG-REPA \citep{agrepa} ranks student layers and timesteps by gate ablation in audio flow matching.
HASTE \citep{haste} stops alignment early, and DyA \citep{dya} and REED \citep{reed} propose other
schedules. Set-level ablations in HASTE and REGLUE \citep{reglue} show that adding shallower
teacher layers hurts, without isolating one layer at a fixed block. Knowledge distillation chooses
teacher layers by curriculum \citep{leap} or unique information \citep{pidkd}, and for language
models the choice matters little \citep{kdlayerselect}. These criteria measure a property of the
teacher, how much existing computation depends on a block, or how quickly a short run improves.
None measures what the student lacks for a specific teacher layer. Section~\ref{sec:criteria}
scores block-level versions of these criteria, similarity measures and gradient signals against the
recoverability gap on the same runs.

\section{Hierarchy Filling}\label{sec:filling}

Let $x$ be an image with latent $z_0$ and noisy latent $z_t$ at noise level $t$. The student is a
SiT \citep{sit} trained by flow matching \citep{flowmatching,rectifiedflow}, with hidden state
$H_{j,t}\in\R^{n\times d_s}$ at block $j$ over $n$ tokens. The teacher is a frozen encoder with
layer-$i$ patch features $T_i(x)\in\R^{n\times d}$, DINOv2-B \citep{dinov2} by default, and $T_0$ is its patch embedding. We read a set $\gI$ of teacher layers and write $i^-$
for the layer before $i$ in $\gI$, with $i^-=0$ for the first; selection uses
$\gI=\{3,6,9,12\}$ and the profiles below use all twelve layers (Appendix~\ref{app:filling}). Every
measurement is in fp32 on a frozen checkpoint of the unaligned run of its setting, fitted and
evaluated on disjoint subsets of a 20K-image calibration split, with noise levels in
three log-SNR bins.

\subsection{Measuring What the Student Has Built}\label{sec:decomposition}

Teacher layers are nested: most of a deep layer is linearly predictable from the layers below it.
Asking whether a student lacks layer~$i$ therefore mixes what the layer adds with everything it
inherits, and cannot say where in the hierarchy the student stops. We measure instead the
\emph{increment} $\Delta T_i$, the part of layer~$i$ that layer~$i^-$ cannot linearly predict. We
whiten every layer to $r=128$ dimensions, $\widetilde{T}_i$, and fit a matrix $B_i\in\R^{r\times r}$
that predicts $\widetilde{T}_i$ from $\widetilde{T}_{i^-}$ by cross-validated reduced-rank ridge
regression. The increment is the residual of this prediction,
\begin{equation}
\Delta T_i = c_i\,\big(\widetilde{T}_i - \widetilde{T}_{i^-}B_i\big)\,D_i^{-1},
\label{eq:decomposition}
\end{equation}
where the diagonal $D_i$ holds the standard deviation of each residual dimension and the scalar
$c_i$ gives $\Delta T_i$ the energy of the whole layer (Appendix~\ref{app:filling}). Whole layers
change little with depth, while each increment carries different content (Figure~\ref{fig:increment}).

To test whether the student has built an increment, we fit a linear map from its hidden state to
the increment and evaluate it on held-out images. The readout $Q_{i,j,t}:\R^{d_s}\to\R^{r}$, one
per teacher layer $i$, block $j$ and noise level $t$, is a ridge regression applied to every token
separately; it only measures and is not the projector used in training. Its score is the held-out
explained variance
\begin{equation}
R^{2}_{i,j,t} = 1 - \frac{\sum_{x}\big\|\Delta T_i(x) - Q_{i,j,t}\big(H_{j,t}(x)\big)\big\|_F^{2}}
                        {\sum_{x}\big\|\Delta T_i(x) - \mathbf{1}\,\overline{\Delta T_i}\big\|_F^{2}},
\label{eq:readout}
\end{equation}
the error relative to predicting the mean token $\overline{\Delta T_i}$, summed over held-out images
$x$ and tokens: $R^2=1$ means the block encodes the increment exactly, $R^2\approx0$ that a linear
map finds none of it. The same readout from the
clean latent $z_0$ gives the availability $A_i$, the share of the increment already linearly
present in the input.

\subsection{The Student Fills the Hierarchy from the Bottom Up}\label{sec:readout}

\begin{figure}[t]
\centering
\fpanel[0.3148\textwidth][0.125in]{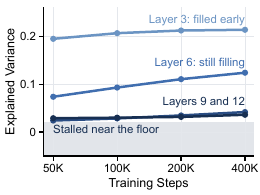}{a}%
\hfill
\fpanel[0.3537\textwidth][0.200in]{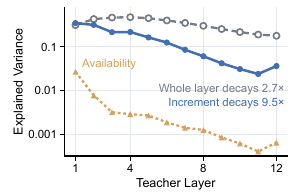}{b}%
\hfill
\fpanel[0.3312\textwidth][0.180in]{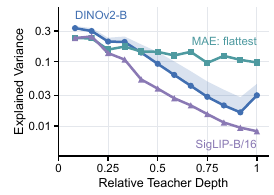}{c}%
\vspace{-5pt}
\caption{\textbf{Hierarchy filling.} The vertical axis is the held-out explained variance $R^2$ of
Eq.~\ref{eq:readout}, the share of a teacher target that a linear readout reconstructs from the
frozen student, maximized over blocks and noise levels on the twelve-layer map. Higher means the
student already encodes more of that target and leaves less for alignment to add. The grey band
lies below $\tau=0.02$, the estimation noise of a held-out $R^2$. \textbf{(a)}~Increments of four
teacher layers over training.
\textbf{(b)}~Whole whitened layer, increment and clean-latent availability at 400K.
\textbf{(c)}~Three teachers at 100K; the shaded region spans SiT-B/2, SiT-L/2, SiT-XL/2 and
Places365 under DINOv2-B.}
\label{fig:filling}
\end{figure}

The student builds shallow increments early and deep ones hardly at all
(Figure~\ref{fig:filling}a). Through 400K steps the best readout of layer 3 stays near 0.2, while
layers 9 and 12 stay below 0.05, close to the readout floor. Whole layers hide this stall, because
a deep layer carries shallow content forward (Figure~\ref{fig:filling}b). The stall is neither a
readout limit, as 0.18 of the whole layer 12 is recovered, nor an input artifact, as availability
is at most 0.03. The profile holds across model scales, on Places365 and under SigLIP, and is
flatter under MAE (Figure~\ref{fig:filling}c; Section~\ref{sec:transfer}). The generative objective
thus builds the shallow increments by itself, consistent with evidence that diffusion models acquire higher-level structure late
\citep{favero2025}, and leaves the top unbuilt within our budgets. That top, where REPA's default
target lies, is where a teacher has something to add.

\section{The Recoverability Gap}\label{sec:gap}

\subsection{From Readout to Gap}\label{sec:gap_def}

Hierarchy filling suggests aligning where the student cannot recover what the teacher adds.
Turning a readout into a gap requires a ceiling, and because selection compares teacher layers the
ceiling must be shared: normalized by its own best readout, a layer the student recovers nowhere
would be scaled by its own estimation noise and look complete. We normalize by the best readout the
measurement achieves anywhere:
\begin{gather}
R^{2}_{\ast} = \max\Big\{\tau,\;\max_{i'\in\gI}A_{i'},\;\max_{i'\in\gI,\,j',\,t'}R^{2}_{i',j',t'}\Big\},
  \label{eq:ceiling} \\[3pt]
N_{i,j,t} = \operatorname{clip}_{[0,1]}\bigg(1 - \frac{R^{2}_{i,j,t}}{R^{2}_{\ast}}\bigg).
  \label{eq:recoverability_gap}
\end{gather}
where $\operatorname{clip}_{[a,b]}(u)=\min\{\max\{u,a\},b\}$. The floor $\tau=0.02$ is the estimation noise of a held-out $R^2$; a layer whose best readout stays
below it cannot be located and is excluded from selection (Appendix~\ref{app:filling}). In terms of
usable information \citep{vinformation}, $N$ is the share of an increment that the student's state
does not carry under a linear readout family. Averaged over noise levels, $N$ forms a map over
teacher layers and student blocks, and selection takes the teacher layer of its argmax.

\subsection{Which Criteria Predict Alignment Benefit?}\label{sec:criteria}

A criterion is useful only if it ranks candidate placements in the order of their benefit. We test
this on fourteen diagnostic branches that fork the unaligned SiT-B/2 at 10K steps and train 50K
further steps, each aligning one whitened target at one block under a shared projector, loss weight
and budget: increments of all four layers at several blocks, plus content and function-class
controls (Appendix~\ref{app:branches}). Benefit is the improvement over the unaligned branch in
\emph{gFID}, an FID on 10K samples with labels and noise shared across branches. Duplicated branches
bound the seed spread at 0.58 gFID, so we treat separations above about 1.5 gFID as reliable. The
published REPA configuration, $+19.87$ gFID, is a positive control. Since a criterion is used to
choose among placements, what matters is the order, not the scale of its values. We
therefore score each criterion by the Spearman rank correlation $\rho$, the standard correlation
between two rankings: $\rho=+1$ if the criterion orders the branches as their
benefit, $-1$ if exactly in reverse, and about $0$ if it carries no ordering information.

\begin{table}[t]
\centering
\caption{\textbf{Placement criteria scored against realized alignment benefit.} Spearman rank
$\rho$ between each criterion, read from the unaligned run at 10K, 30K and 50K steps,
and the gFID benefit of the fourteen branches of Section~\ref{sec:criteria}; tied values share their
mean rank. Stable marks a criterion whose percentile-bootstrap 95\% interval excludes zero at all
three checkpoints, in either direction.}
\label{tab:criteria}
\tablestyle
\renewcommand{\arraystretch}{1.11}
\setlength{\tabcolsep}{3pt}
\begin{tabular}{ll*{3}{w{c}{25.5pt}}c}
\toprule
\vhd{Criterion} & \vhd{What It Measures} & \multicolumn{3}{c}{\hd{Spearman $\rho$ with Benefit}} & \vhd{Stable} \\
\cmidrule(lr){3-5}
 & & \hd{10K} & \hd{30K} & \hd{50K} & \\
\midrule
\rowcolor{oursrow}
\textbf{Recoverability gap $N$ (Ours)} & Unrecovered share of the target & $+0.86$ & $+0.90$ & $+0.92$ & \cmark \\
\grouprule{6}
\multicolumn{6}{l}{\textit{Similarity to the Target}} \\
Readout $R^2$ & Recovered share of the target & $-0.85$ & $-0.87$ & $-0.89$ & \cmark \\
CKA & Similarity to the target & $-0.81$ & $-0.84$ & $-0.85$ & \cmark \\
Gram similarity & Match of token-to-token relations & $+0.37$ & $-0.37$ & $-0.27$ & \xmark \\
\grouprule{6}
\multicolumn{6}{l}{\textit{Property of the Student Block}} \\
LDS (iREPA) & Spatial structure of the block & $+0.15$ & $-0.03$ & $-0.40$ & \xmark \\
Linear probing (REPA) & Class separability of the block & $-0.40$ & $-0.01$ & $-0.11$ & \xmark \\
Gate ablation (AG-REPA) & Loss increase if the block is removed & $+0.31$ & $+0.31$ & $+0.43$ & \xmark \\
\grouprule{6}
\multicolumn{6}{l}{\textit{First-Order Signal}} \\
Gradient cosine (HASTE) & Alignment--flow gradient agreement & $-0.91$ & $-0.93$ & $-0.49$ & \cmark \\
First-order utility $U$ & Flow-loss slope toward the target & $-0.34$ & $+0.00$ & $-0.04$ & \xmark \\
$N\cdot[U]_+$ (fixed in advance) & Gap times positive slope & $-0.34$ & $-0.28$ & $-0.25$ & \xmark \\
\bottomrule
\end{tabular}

\end{table}

\begin{figure}[t]
\centering
\scalebox{0.96}{%
\begin{minipage}{\textwidth}
\fpanel[0.3424\textwidth][0.125in]{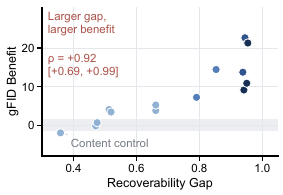}{a}%
\hfill
\fpanel[0.3515\textwidth][0.150in]{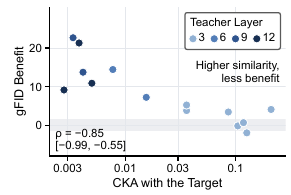}{b}%
\hfill
\fpanel[0.3057\textwidth][0.150in]{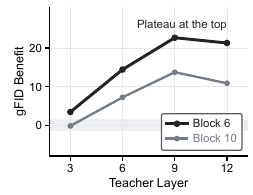}{c}%
\end{minipage}}
\vspace{-5pt}
\caption{\textbf{The gap predicts alignment benefit; similarity predicts it in reverse.}
\textbf{(a,\,b)}~Benefit of the fourteen branches against the gap and against CKA, on a logarithmic
axis, on the 50K map, colored by teacher layer, with Spearman $\rho$ and its 95\% interval; the grey band marks benefits
within 1.5 gFID of zero. \textbf{(c)}~Benefit along the teacher axis at blocks 6 and 10.}
\vspace{0pt}
\label{fig:criteria}
\end{figure}

Only the gap ranks placements stably in the right direction
(Table~\ref{tab:criteria}, Figure~\ref{fig:criteria}a), and the shared ceiling carries this signal:
with each layer normalized by its own best readout, the same quantity ranks placements at $-0.39$.
Readout $R^2$ mirrors the gap by construction; the independent evidence is CKA \citep{cka}, which
ranks placements in reverse (Figure~\ref{fig:criteria}b), as hierarchy filling predicts: a block
resembles the teacher where the objective has filled the hierarchy. The gradient cosine is
negative for the same reason, since the two gradients agree most where the flow objective already
moves the student toward the target. Properties of the student block keep no stable sign, and
neither does the short-run signal, the branches' own held-out flow loss ($\rho=-0.04$).

The gap orders placements along the teacher axis (Figure~\ref{fig:criteria}c), and two simpler
readings of this order fail. If any clean, sample-specific target helped as a regularizer, targets
of equal dimension and energy at the same block would not spread over 19 gFID, and the content
control, the lowest-gap target, would not degrade generation by 2.0 gFID. Nor is the gap a proxy
for depth: within layer 3, where depth is fixed, it still orders the seven branches
($\rho=+0.86$, 95\% interval $[0.32,1]$), and at the top, layer 9 beats layer 12 on both seeds, by
1.4 and 1.8 gFID. The gap does not resolve that last difference: the two layers differ by 0.009 in
gap, less than the 0.025 spread of a single cell across noise levels. Its argmax, layer 12 at block
6, thus lands on the best plateau rather than the best cell, and both plateau cells exceed the
published REPA configuration ($+21.3$ and $+22.7$ against $+19.9$ gFID). The same ordering explains
why adding shallower teacher layers hurts \citep{haste,reglue}.

The gap does not choose the block. For layer 12 it varies by less than 0.02 across blocks, while
benefit more than doubles from block 2 to block 6 (Figure~\ref{fig:map}b): the gap tells whether
the student has a layer's content, not where to supply it.

\textbf{Distance, not slope.} An alternative to the distance to a layer is the slope toward it, the
logic of HASTE's gradient angle. Its finest-grained form, the first-order utility $U$, the relative
flow-loss reduction per unit step of the hidden state toward the target
(Appendix~\ref{app:utility}), does not rank placements, and the product $N\cdot[U]_+$, fixed in
advance, erases the signal of $N$ (Table~\ref{tab:criteria}); four controls attribute this to $U$
rather than to its estimator. Benefit over many steps follows the distance to the target, not the
local slope: a slope may signal when to stop aligning, as in HASTE, but not where to
align.

\section{RARE: Aligning Where the Gap Is}\label{sec:rare}

RARE takes the teacher layer from the gap, follows the gap's training-time counterpart, the
alignment residual, across tokens and training steps, and keeps the student block of the baselines.

\textbf{Target.} RARE aligns the teacher layer with the largest gap, $i^\star$, fixed from the map
before alignment. On ImageNet with SiT-B/2 this is layer 12, the layer REPA and iREPA align
by default; on Places365 it is layer 9. The training target is the whitened layer
$\widetilde{T}_{i^\star}$ rather than its increment. The increment locates the stall, but layers 9
and 12 form one plateau of the gap (Section~\ref{sec:criteria}), and the whitened layer carries
both increments. It beats the increment in every setting we test
(Tables~\ref{tab:ablation} and~\ref{tab:transfer_inc}).

\textbf{Depth.} Since the gap does not separate blocks, RARE inserts the target at the block of the
released REPA and iREPA configurations, $j_{\mathrm{b}}=4$ on SiT-B/2 and $j_{\mathrm{b}}=8$ on
SiT-L/2 and SiT-XL/2, so every comparison with them is placement-matched. Block 6, where the
selected cell lies, is an ablation.

\textbf{Release.} The gap is read on frozen checkpoints. During training RARE tracks the alignment
residual defined below and releases the loss once it stops falling: past that point alignment
constrains the student without closing the gap further, the capacity cost HASTE identifies.

During alignment the loss is
\begin{equation}
\Ls = \Ls_{\mathrm{flow}} - \lambda(s)\,\frac{1}{n}\sum_{k=1}^{n} w_k\cos_k,
\qquad
\cos_k = \cos\big(P(H_{j_{\mathrm{b}},t})_k,\;\operatorname{sn}(\widetilde{T}_{i^\star})_k\big),
\label{eq:objective}
\end{equation}
the sparse alignment loss (SPAlign in Figure~\ref{fig:teaser}), sparse because it acts on one
teacher layer, concentrates on the tokens that still miss their target, and is active only until
the release. Here $s$ is the training step, $P$
is the convolutional projector of iREPA and $\operatorname{sn}$ its spatial normalization
(Appendix~\ref{app:impl}); the target uses the frozen calibration statistics
$(W_{i^\star},\mu_{i^\star})$, and the projector is discarded after training. Let
$\eta_k=\tfrac{1}{2}(1-\cos_k)$ be the alignment residual of token $k$, the part of the target its
projected state still misses. The token weights follow the residual:
\begin{equation}
\tilde{w}_k = \operatorname{clip}_{[w_{\min},\,w_{\max}]}\bigg(\frac{\mathrm{sg}(\eta_k)}{\frac{1}{n}\sum_{k'}\mathrm{sg}(\eta_{k'})}\bigg),
\qquad
w_k = \frac{\tilde{w}_k}{\frac{1}{n}\sum_{k'}\tilde{w}_{k'}},
\label{eq:tokenweight}
\end{equation}
where $\mathrm{sg}$ is the stop-gradient, so the weights only reallocate the alignment gradient across
tokens. A token's raw weight is its residual relative to the mean residual; the bounds
$w_{\min}=1/4$ and $w_{\max}=4$ are hyperparameters that keep a token that already matches the target
in the loss and stop a few outlier tokens from dominating it. The schedule stops
spending once the residual plateaus. Let $\hat{g}_s$ and $\bar{g}_s$ be exponential moving
averages of the mean residual with decays $0.999$ and $0.99995$, and let $s_0$ be the first step
after a warm-up of $S$ steps at which the fast average stops improving on the slow one,
$(\bar{g}_s-\hat{g}_s)/\bar{g}_s<\delta=2\%$. With the cosine ramp
$\phi(s;a)=\tfrac{1}{2}\big(1+\cos(\pi\operatorname{clip}_{[0,1]}\tfrac{s-a}{S})\big)$, where $S$
is one eighth of the training budget ($50\mathrm{K}$ steps in 400K-step runs), and $\phi(s;s_0)=1$
before the plateau is detected,
\begin{equation}
\lambda(s) = \lambda_0\,\min\big\{\phi(s;s_0),\;\phi(s;s_{\mathrm{h}}-S)\big\},
\qquad \lambda_0 = 1,\quad s_{\mathrm{h}} = 250\mathrm{K},
\label{eq:schedule}
\end{equation}
so the loss reaches zero within $S$ steps of the plateau, and by the 250K horizon of HASTE at the
latest. On ImageNet the plateau is detected at step 104K and the loss reaches zero at 154K.

\section{Experiments}\label{sec:experiments}

In this section, we conduct experiments to answer the following research questions:
\begin{itemize}[leftmargin=*,itemsep=1pt,topsep=2pt]
\item \textbf{RQ1:} Does placing alignment by the gap improve on existing methods on ImageNet-256?
\item \textbf{RQ2:} How much do the target, block, token weights and release of RARE each contribute?
\item \textbf{RQ3:} How does alignment change training speed and the gap at the layer it targets?
\item \textbf{RQ4:} Does the recipe transfer across model scale, teacher, image domain and backbone?
\item \textbf{RQ5:} What do selecting the layer and training with RARE cost?
\end{itemize}

\begin{table}[h]
\centering
\setlength{\abovecaptionskip}{3pt}%
\caption{\textbf{ImageNet-256 with SiT-B/2 at 400K steps.} L12 is DINOv2-B's last layer, aligned
raw (768 dimensions) unless marked whitened (128 dimensions); attn.\ is HASTE's attention-map term.}
\label{tab:imagenet}
\tablestyle
\renewcommand{\arraystretch}{1.15}
\setlength{\tabcolsep}{2.1pt}
\begin{tabular}{lllrrrrrr}
\toprule
\vhd{Method} & \multicolumn{2}{c}{\hd{Alignment}} & \vhd{FID\down} & \vhd{sFID\down} & \vhd{IS\up} & \vhd{Prec.\up} & \vhd{Rec.\up} & \vhd{CMMD\down} \\
\cmidrule(lr){2-3}
 & \hd{Target} & \hd{Block} & & & & & & \\
\midrule
\grouphead{9}{Without Guidance}
\rowcolor{baserow}
SiT~\citep{sit}\venue{ECCV'24} & --- & --- & 35.15 & 6.60 & 41.94 & 0.526 & 0.633 & 1.279 \\
\grouprule{9}
REPA~\citep{repa}\venue{ICLR'25} & L12 & 4 & 22.39 & 6.62 & 65.62 & 0.594 & \underline{0.648} & 1.067 \\
iREPA~\citep{irepa}\venue{ICLR'26} & L12 & 4 & 20.40 & 6.90 & 72.61 & 0.602 & \textbf{0.650} & 1.054 \\
HASTE~\citep{haste}\venue{NeurIPS'25} & L12 + attn. & 8 & \underline{18.99} & 6.52 & \underline{74.13} & \underline{0.624} & 0.638 & \underline{0.984} \\
\grouprule{9}
sREPA~\citep{srepa}\venue{arXiv'26} & L12 + Gram & 4 & 21.64 & 6.77 & 68.83 & 0.596 & 0.645 & 1.060 \\
SRA~\citep{sra}\venue{ICLR'26} & Self (EMA) & 4 & 28.84 & \textbf{6.15} & 51.35 & 0.563 & 0.645 & 1.146 \\
AHPA~\citep{ahpa}\venue{arXiv'26} & VAE prior & 3 & 37.52 & 6.81 & 40.09 & 0.510 & 0.637 & 1.327 \\
SPARE~\citep{spare}\venue{arXiv'26} & Latent affinity & 4 & 31.36 & \underline{6.35} & 46.81 & 0.548 & 0.643 & 1.202 \\
\midrule
\rowcolor{oursrow}
\textbf{RARE (Ours)} & L12, whitened & 4 & \textbf{18.02} & \underline{6.35} & \textbf{77.44} & \textbf{0.626} & 0.640 & \textbf{0.970} \\
\midrule
\grouphead{9}{With Guidance (scale 1.65 on [0,\,0.72])}
\rowcolor{baserow}
SiT & --- & --- & 14.33 & 5.23 & 96.93 & 0.689 & 0.559 & 0.994 \\
\grouprule{9}
REPA & L12 & 4 & 6.69 & 5.26 & 155.45 & 0.737 & \textbf{0.590} & 0.828 \\
iREPA & L12 & 4 & 5.48 & 5.39 & 172.89 & 0.748 & 0.588 & 0.809 \\
HASTE & L12 + attn. & 8 & \underline{5.00} & 5.28 & \underline{176.12} & \underline{0.767} & 0.577 & \underline{0.760} \\
\grouprule{9}
sREPA & L12 + Gram & 4 & 6.12 & 5.34 & 162.26 & 0.742 & \underline{0.589} & 0.819 \\
SRA & Self (EMA) & 4 & 10.47 & \textbf{4.94} & 116.29 & 0.726 & 0.554 & 0.883 \\
AHPA & VAE prior & 3 & 15.84 & 5.40 & 91.27 & 0.674 & 0.565 & 1.038 \\
SPARE & Latent affinity & 4 & 11.97 & \underline{5.07} & 108.62 & 0.710 & 0.556 & 0.931 \\
\midrule
\rowcolor{oursrow}
\textbf{RARE (Ours)} & L12, whitened & 4 & \textbf{4.46} & 5.09 & \textbf{182.00} & \textbf{0.781} & 0.566 & \textbf{0.732} \\
\bottomrule
\end{tabular}
\vspace{-5pt}
\end{table}

\subsection{Experimental Setup}\label{sec:setup}

\textbf{Data and models.} ImageNet-1K at $256\times256$ \citep{imagenet} is the main benchmark,
and Places365-Standard \citep{places365} changes the image distribution under the same
class-conditional pipeline. The student is SiT-B/2, with SiT-L/2 and SiT-XL/2 for the scale study,
and the tokenizer is the SD-VAE of \citet{ldm}. The teachers are DINOv2-B, MAE-B/16 \citep{mae}
and SigLIP-B/16 \citep{siglip}, three ViT-B encoders with twelve layers and the same $16\times16$
patch grid, pre-trained by self-distillation, pixel reconstruction and language supervision
\citep{clip}.

\textbf{Baselines.} The baselines are the unaligned SiT, REPA, iREPA and HASTE in their published
configurations, and four objectives that change what is aligned: token-relation structure (sREPA),
the model's own deeper EMA layer (SRA), a frozen VAE prior with a timestep router (AHPA), and a
latent affinity without any encoder (SPARE, \citealt{spare}).

\textbf{Training and evaluation.} All runs share one initialization and seed and train with batch
size 256, AdamW at a constant learning rate of $10^{-4}$, fp16 and EMA decay 0.9999 on eight V100
GPUs. ImageNet and Places365 runs train for 400K steps, about 80 epochs; the scale, teacher and
MM-DiT studies stop at a 100K endpoint fixed in advance. We follow the ADM protocol \citep{adm}:
50K samples from a 250-step Euler--Maruyama SDE sampler, scored by FID \citep{fid}, sFID, Inception
Score \citep{inceptionscore}, precision and recall \citep{precisionrecall}, and CMMD \citep{cmmd},
which does not use Inception features; Places365 adds KID \citep{kid}. All methods share class
labels, initial noise and sampler noise. Unguided results are the main comparison, and at the
endpoint we also report classifier-free guidance (CFG) of scale 1.65 on the interval $[0,0.72]$.
Sample figures keep class and noise fixed within each column and use guidance 4.0 for
visualization. Unless a table says otherwise, the setting is SiT-B/2 with DINOv2-B on
ImageNet-256, the best value is in bold and the second best underlined. With one training seed per
run, FID differences below about 1.5, the separation the seed spread of Section~\ref{sec:criteria}
supports, are read for their direction only.

\subsection{Comparison on ImageNet-256 (RQ1--RQ3)}\label{sec:imagenet}

\begin{figure}[t]
    \centering
    \scalebox{0.96}{%
    \begin{minipage}{\textwidth}
    \fpanel[0.3611\textwidth][0.140in]{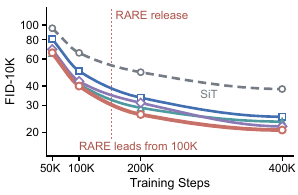}{a}%
    \hfill
    \fpanel[0.3720\textwidth][0.170in]{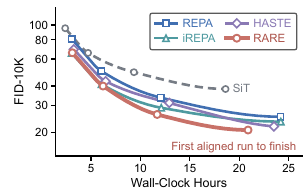}{b}%
    \hfill
    \fpanel[0.2666\textwidth][0.160in]{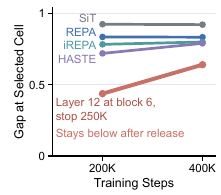}{c}%
    \end{minipage}}
    \vspace{-8pt}
    \caption{\textbf{Training dynamics.} \textbf{(a,\,b)}~FID-10K without guidance on a logarithmic
    axis against steps, mapped by $(s/100\mathrm{K})^{1.2}$ to spread the early checkpoints, and against
    wall-clock hours; markers are measured checkpoints. \textbf{(c)}~Gap at the selected cell (layer
    12, block 6; twelve-layer map) at 200K and 400K, including the run that aligns this cell until
    250K (Table~\ref{tab:ablation}, row 4).}
    \label{fig:training}
    \vspace{-1.5pt}
    \end{figure}

\textbf{Main comparison (RQ1).} RARE has the best FID, IS, precision and CMMD with and without guidance (Table~\ref{tab:imagenet}),
ahead of iREPA by 2.38 FID without guidance. The contrast between the baseline groups is as
informative: the three methods that replace the pretrained teacher with a signal already in the
system, SRA, SPARE and AHPA, trail every teacher-based method, and AHPA does not improve on the
unaligned SiT, as expected if alignment supplies content the student lacks.

\begin{table}[!htbp]
\setlength{\abovecaptionskip}{0pt}%
\setlength{\belowcaptionskip}{6pt}%
\centering
\caption{\textbf{Ablation of RARE's decisions} at 400K steps. All rows align layer 12, whitened or
as its increment, with RARE's projector and loss weight. $w_k$: token weights (Eq.~\ref{eq:tokenweight});
Measured: release of Eq.~\ref{eq:schedule}; Ramp: hand-set. Row 7 is RARE.}
\label{tab:ablation}
\small
\setlength{\aboverulesep}{0pt}%
\setlength{\belowrulesep}{0pt}%
\renewcommand{\arraystretch}{1.3}%
\setlength{\tabcolsep}{5pt}
\begin{tabular}{clcclrr}
\toprule
\vhd{\#} & \vhd{Target} & \vhd{Block} & \vhd{$w_k$} & \vhd{Schedule} & \multicolumn{2}{c}{\hd{FID-50K\down}} \\
\cmidrule(lr){6-7}
 & & & & & \hd{no CFG} & \hd{CFG} \\
\midrule
1 & Whitened & 4 & \xmark & Always on & 20.40 & 5.48 \\
2 & Whitened & 6 & \xmark & Always on & 20.86 & 5.85 \\
3 & Increment & 6 & \xmark & Always on & 21.79 & 6.42 \\
\grouprule{7}
4 & Whitened & 6 & \xmark & Stop at 250K & 19.32 & 4.90 \\
5 & Whitened & 4 & \xmark & Stop at 250K & 18.83 & 4.77 \\
\grouprule{7}
6 & Whitened & 4 & \cmark & Ramp 200K$\to$250K & \textbf{17.85} & \textbf{4.45} \\
\rowcolor{oursrow}
7 & Whitened & 4 & \cmark & Measured & \underline{18.02} & \underline{4.46} \\
\bottomrule
\end{tabular}
\end{table}
\textbf{Ablation (RQ2).} Table~\ref{tab:ablation} changes one decision at a time, except rows 5 and 6, which change the token
weights and the release together. Two results carry the design. The 128-dimensional
whitened layer ties iREPA's 768-dimensional target (row 1), so one sixth of the dimensions keeps
what alignment needs. Releasing the loss is the largest single gain, 1.54 FID over keeping it on
(rows 2 and 4). Block 6 and the increment target each raise FID (rows 1 to 3), and
residual-proportional weights with a ramped release lower it (rows 5 and 6). The measured release
trails the hand-set ramp by 0.17 FID at 400K but stops 96K steps earlier and leads by 2.97 FID-10K
at 200K (rows 6 and 7).

\textbf{Training dynamics (RQ3).} RARE also leads during training (Figure~\ref{fig:training}a,b): it has the lowest FID-10K from
100K steps on and finishes 400K steps in 20.9 hours against 24.2 for iREPA. The gap at the selected
cell shows what alignment does to the student (Figure~\ref{fig:training}c). The run that aligns
this cell lowers its gap from 0.92 to 0.44 by 200K steps, below the 0.72 to 0.83 of REPA, iREPA and
HASTE, which align layer 12 at blocks 4 and 8. At 400K, 150K steps after the release, the gap has
risen only to 0.64: releasing the loss keeps most of what alignment filled. The closure concentrates
on the aligned layer, 0.49 of the gap at layer 12 against at most 0.27 elsewhere on the map
(Figure~\ref{fig:closure_map}). In samples, only RARE renders the peacock's open train at 400K
(Figure~\ref{fig:qualitative}), and along the sampler the aligned models fix the object earlier
rather than sharpening the last steps (Figure~\ref{fig:trajectory}).

\subsection{Transfer Across Scale, Teacher and Domain (RQ4)}\label{sec:transfer}

\setlength{\intextsep}{0pt}%
\begin{wraptable}[16]{r}{0.46\textwidth}
\begin{minipage}{\linewidth}
\setlength{\abovecaptionskip}{0pt}%
\centering
\caption{\textbf{Transfer.} Unguided FID-50K except the CFG row; RARE aligns layer 9 on Places365.
Bold: better of iREPA and RARE.}
\label{tab:transfer}
\footnotesize
\setlength{\aboverulesep}{0pt}%
\setlength{\belowrulesep}{0pt}%
\renewcommand{\arraystretch}{1.15}%
\setlength{\tabcolsep}{3pt}
\begin{tabular}{lrrr}
\toprule
\vhd{Setting} & \multicolumn{3}{c}{\hd{FID\down}} \\
\cmidrule(lr){2-4}
 & \hd{SiT} & \hd{iREPA} & \hd{RARE} \\
\midrule
\grouphead{4}{(a) Scale, DINOv2-B, 100K Steps}
SiT-B/2 & 62.59 & 37.70 & \textbf{36.66} \\
SiT-L/2 & 47.19 & 19.94 & \textbf{18.93} \\
SiT-XL/2 & 42.45 & 16.52 & \textbf{16.07} \\
\midrule
\grouphead{4}{(b) Teacher, SiT-B/2, 100K Steps}
MAE-B/16 & 62.59 & 49.95 & \textbf{43.81} \\
SigLIP-B/16 & 62.59 & 40.60 & \textbf{39.86} \\
\midrule
\grouphead{4}{(c) Places365, SiT-B/2, 400K Steps}
no CFG & 11.40 & 7.71 & \textbf{7.33} \\
CFG & 6.79 & 4.98 & \textbf{4.54} \\
\bottomrule
\end{tabular}
\end{minipage}
\end{wraptable}
ImageNet runs keep the recipe of SiT-B/2, and on Places365 RARE aligns layer 9, the layer its map
selects. RARE has a lower FID than iREPA in all six settings of Table~\ref{tab:transfer}, most
clearly with MAE (6.14 FID); Table~\ref{tab:transfer_full} reports all metrics and
Figure~\ref{fig:places} shows Places365 samples. The whitened layer also beats the increment in
every setting, by 1.86 to 5.33 FID, even when the increment is aligned at the cell its own map
selects (Table~\ref{tab:transfer_inc}).

\textbf{Other backbones.} The ordering holds when the generative objective or the architecture
changes. On DiT-B/2 with $\epsilon$-prediction and a DDPM sampler, and on MM-DiT-B/2, whose image
and class tokens keep separate weights and share one attention, RARE has the lowest FID with and
without guidance (Tables~\ref{tab:dit} and~\ref{tab:mmdit}); without guidance it reaches 20.54 and
32.98 against 21.76 and 34.28 for HASTE, the strongest baseline.

\textbf{Teacher.} The teacher comparison agrees with a prediction of hierarchy
filling: a teacher whose hierarchy the student fills more completely leaves less for alignment to
add. Under MAE the largest gap along the teacher axis is 0.82, against 0.96 for DINOv2 and 0.95 for
SigLIP, and MAE gives the smallest improvement over the unaligned SiT, 18.78 FID against 25.93 and
22.73.

\setlength{\intextsep}{8pt plus 2pt minus 2pt}%
\subsection{Training and Selection Cost (RQ5)}\label{sec:efficiency}

RARE trains at 0.188 seconds per step against 0.218 for iREPA and reaches 400K steps in 167
GPU-hours, 14\% fewer than iREPA and 11\% fewer than HASTE, because its target has 128 dimensions
and its loss is released before the midpoint (Table~\ref{tab:cost}). Selecting the layer costs one
map of a 50K-step unaligned checkpoint, 0.15 to 0.35 GPU-hours, plus 19 GPU-hours to train that
checkpoint if it is not already available. Training every candidate cell to 100K steps instead
costs 836 GPU-hours, and its best cell leads the map's choice by 1.37 gFID (Figure~\ref{fig:cost}).

\section{Conclusion, Limitations and Future Work}

Representation alignment helps where the student cannot linearly recover the teacher's features, not
where it already resembles them. The generative objective fills the hierarchy bottom-up and stalls
near the top; the gap it leaves ranks placements by benefit. RARE takes its target from the gap and
its schedule from the alignment residual, and beats REPA, iREPA and HASTE at lower training cost.

The four-layer map does not separate its top two layers, so its argmax trails the best cell by 1.4
to 1.8 gFID, and it neither chooses the student block nor predicts the size of a benefit. The
evidence covers SiT-B/L/XL, DiT-B/2 and MM-DiT-B/2 with three ViT-B teachers on ImageNet and
Places365, one seed per setting and 100K endpoints in the scale, teacher and MM-DiT studies. The
same measurement could guide other models that inject a pretrained representation, such as
tokenizers on frozen encoders and unified understanding-and-generation models, before any aligned
run is trained.

\clearpage
\section*{AI use statement}
We used generative AI tools to assist with figure preparation and to edit and proofread the
manuscript.

\section*{Ethics statement}
This work trains class-conditional image generators on ImageNet-1K and Places365-Standard and
measures them against publicly released pretrained encoders. Both datasets are standard public
research corpora; no human subjects were involved and no new data were collected. The method
lowers the training cost of an existing class of generative models without extending what they can
produce, so it inherits the known risks of image synthesis rather than adding one.

\section*{Reproducibility statement}
Section~\ref{sec:setup} gives the datasets, backbones, tokenizer, optimizer, batch size, precision,
hardware, budget and the paired sampling protocol under which every number is scored.
Sections~\ref{sec:filling} to~\ref{sec:rare} and Appendix~\ref{app:filling} define the decomposition, the recoverability gap and
the alignment loss with the constants they use: the rank $r=128$, the variance floor $\gamma=0.1$,
the readout floor $\tau=0.02$, the token-weight bounds $w_{\min}=1/4$ and $w_{\max}=4$, the
plateau tolerance $\delta=2\%$, the loss weight $\lambda_0=1$
and the release ramp of one eighth of the training budget. Every map is computed in fp32 from one frozen checkpoint of the
unaligned baseline of its own setting, on a 20K-image calibration split with disjoint fit,
development and evaluation subsets. Appendices~\ref{app:filling} to~\ref{app:impl} give the decomposition constants, the
diagnostic branches, the estimation protocol and controls of the first-order utility, and the
projector of RARE.


\bibliography{main}
\bibliographystyle{iclr2027_conference}

\clearpage
\appendix
\makeatletter
\setlength{\@fpsep}{12pt plus 2pt minus 2pt}
\setlength{\@fpbot}{0pt plus 1fil}
\newcommand{\appendixpart}[2]{%
  \FloatBarrier
  \par\addpenalty{-300}\addvspace{10pt plus 3pt minus 2pt}%
  \pdfbookmark[0]{Part #1: #2}{appendixpart.#1}%
  \nobreak\hrule height 1pt\nobreak\vskip 4pt\nobreak
  \hbox to\textwidth{\large\sc Part~#1\hspace{0.8em}#2\hfil}\nobreak
  \vskip 4pt\nobreak\hrule height 0.4pt\nobreak\vskip 2pt\nobreak
  \@nobreaktrue}
\let\appendixsection\section
\renewcommand{\section}{\FloatBarrier\appendixsection}
\makeatother

\appendixpart{I}{The Recoverability Gap}

\section{Measurement Details and Visualization}\label{app:filling}

This section gives the measurement details of Sections~\ref{sec:filling} and~\ref{sec:gap}: the
layer sets, the constants of Eq.~\ref{eq:decomposition}, the readout floor and ceiling, and a view
of the increments.

\textbf{Layer sets.} Selection and criterion scoring read $\gI=\{3,6,9,12\}$. The profiles of
Figure~\ref{fig:filling}, the map of Figure~\ref{fig:map} and the gaps of aligned runs read all
twelve layers; their increments are defined against different predecessors and are never compared
with the four-layer ones. The calibration split has disjoint fit, development and evaluation
subsets, and the measurement never updates the student.

\textbf{Decomposition.} Each layer is whitened to $r=128$ dimensions with its own mean $\mu_i$ and
matrix $W_i$, estimated on the fit subset, so that every layer enters the readout with the same
dimension and unit variance per direction:
\begin{equation}
\widetilde{T}_i = \big(T_i - \mathbf{1}\mu_i^{\top}\big)W_i\in\R^{n\times r}.
\label{eq:whitening}
\end{equation}
With $\widehat{T}_i=\widetilde{T}_{i^-}B_i$ the prediction of layer~$i$ from layer~$i^-$, its
inherited part, the two normalizers of Eq.~\ref{eq:decomposition} are
\begin{equation}
D_i = \operatorname{diag}\big(\max(\sigma_{i,1},\gamma),\,\ldots,\,\max(\sigma_{i,r},\gamma)\big),
\qquad
c_i = \frac{\|\widetilde{T}_i\|_F}{\big\|(\widetilde{T}_i-\widehat{T}_i)D_i^{-1}\big\|_F},
\label{eq:decomposition_full}
\end{equation}
where $\sigma_{i,d}$ is the residual standard deviation of dimension $d$. The variance floor
$\gamma=0.1$ keeps residual dimensions that the predecessor almost fully explains from inflating
regression noise. The scale $c_i$ gives every increment the energy of its whole layer; it leaves
$R^2$ unchanged but makes increments and whole layers energy-matched targets when they are aligned
in Section~\ref{sec:criteria}.

\textbf{Readout and ceiling.} The readouts $Q_{i,j,t}$ of Eq.~\ref{eq:readout} use one ridge
strength throughout; the sums run over held-out images and noise draws. Including the availability
in the ceiling of Eq.~\ref{eq:ceiling} keeps it at least at what the clean latent offers; on our maps
it never binds. The null of a held-out $R^2$ lies near zero, and $\tau=0.02$ is its estimation
noise. A layer whose best readout stays below $\tau$ is flagged absent: its gap is 1 when criteria
are scored, and it is excluded from selection. The conclusions of Section~\ref{sec:criteria} are
unchanged for $\tau\in\{0.01,0.02\}$.

\begin{figure}[htbp]
\centering
\includegraphics[width=0.95\textwidth]{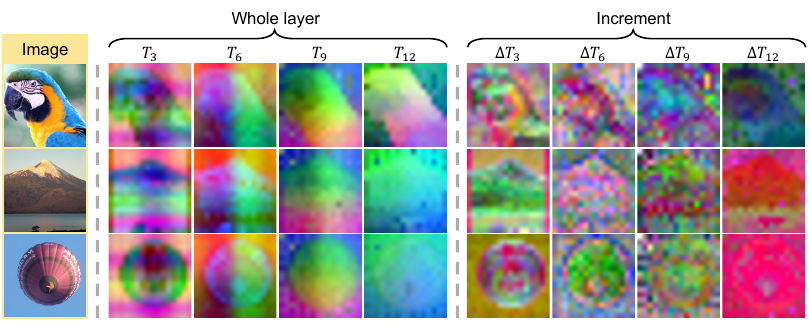}
\caption{\textbf{Whole teacher layers are near-copies of one another; their increments are not.}
Top three principal components of each feature map as RGB, fitted jointly in a shared rank-128
basis. \textbf{Left:} the whole whitened layer separates object from background at every depth and
changes mainly in hue. \textbf{Right:} the increments carry different content at each depth, from
edges and texture at layer 3 to coarse, low-frequency regions at layer 12.}
\label{fig:increment}
\end{figure}

\textbf{Visualization.}
Figure~\ref{fig:increment} renders the whole whitened layers and their increments for three
images: removing what a layer inherits from the layer below leaves qualitatively different content
at each depth, which the readouts of Section~\ref{sec:readout} measure one layer at a time.

\section{Diagnostic Branches}\label{app:branches}

The fourteen branches behind Table~\ref{tab:criteria} and Figure~\ref{fig:criteria} each fork the
unaligned SiT-B/2 at 10K steps and train 50K further steps with one whitened target at one block.
They align the increments of
layers 6 and 9 at blocks 6 and 10, of layer 12 at blocks 2, 6 and 10, and of layer 3 at blocks 4, 6,
8 and 10. At layer 3 they also align the whole whitened layer, the inherited part $\widehat{T}_3$
as a content control, and the increment through a $1\times1$ projector as a function-class
control. Each branch is scored on the map of its own target form. Three branches duplicated with a
second training seed under paired sampling differ by 0.022, 0.158 and 0.580 gFID, which sets the
1.5 gFID separation used throughout.

\section{The Recoverability Gap Map and Its Two Axes}\label{app:map}

\begin{figure}[htbp]
\centering
\fpanel[0.4776\textwidth][-0.070in]{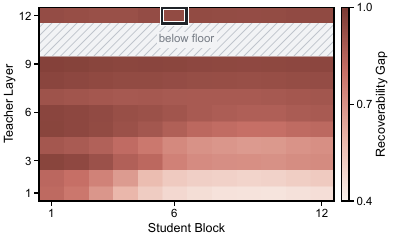}{a}%
\hfill
\fpanel[0.5222\textwidth][0.035in]{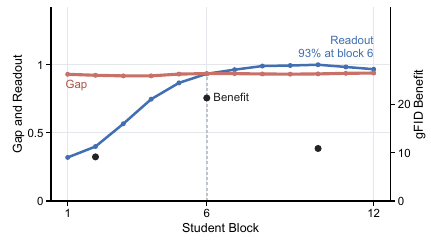}{b}%
\par\vspace{6pt}
\fpanel[0.9999\textwidth][0.325in]{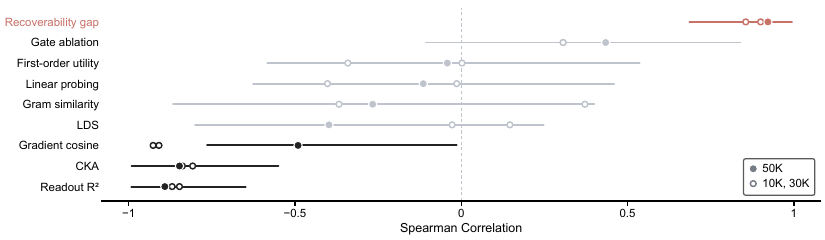}{c}%
\caption{\textbf{The recoverability gap map and its two axes.} \textbf{(a)}~Gap of the unaligned
SiT-B/2 at 50K on all twelve teacher layers; hatched layers fall below the readout floor, and the
box marks the cell the four-layer map selects. \textbf{(b)}~Along the block axis, the gap of layer
12, the readout of the shallowest increment, and the benefit of aligning layer 12 at blocks 2, 6
and 10. \textbf{(c)}~Spearman $\rho$ of the nine single criteria at 50K with bootstrap 95\%
intervals, and at 10K and 30K.}
\label{fig:map}
\end{figure}

Section~\ref{sec:criteria} reads the map along two axes. Along the teacher axis the gap rises with
depth; at block 6 the four-layer gaps of layers 3, 6, 9 and 12 are 0.52, 0.85, 0.94 and 0.95,
against benefits of $+3.4$, $+14.4$, $+22.7$ and $+21.3$ gFID. Along the block axis the gap of
layer 12 stays within 0.02, while aligning it at blocks 2, 6 and 10 gains 9.1, 21.3 and 10.9 gFID
(Figure~\ref{fig:map}b), which is why Section~\ref{sec:rare} takes the insertion block from the
baselines. Figure~\ref{fig:map}a shows the full twelve-layer map of the unaligned SiT-B/2 at 50K
steps. This map defines each increment against the adjacent layer, so its cells are not compared
with the four-layer map used for selection. Figure~\ref{fig:map}c repeats the correlations of
Table~\ref{tab:criteria} with their bootstrap intervals.

\section{First-Order Utility: Definition, Protocol and Controls}\label{app:utility}

Section~\ref{sec:criteria} tests a slope-based criterion, the first-order utility $U$, against the
gap. This section defines $U$, describes its estimator and lists the four controls cited there.

\textbf{Definition.} Let $\Ls_{\mathrm{align}}(H)=\|Q_{i,j,t}(H)-\Delta T_i\|_F^{2}$, let $\hat d$ be
$-\nabla_H\Ls_{\mathrm{align}}$ at $H_{j,t}$ scaled to unit root mean square, and let
$\Ls_{\mathrm{flow}}(H)$ be the flow loss when $H$ replaces the hidden state at block $j$ of the
frozen student. The first-order utility
\begin{equation}
U_{i,j,t} = -\frac{\operatorname{RMS}(H_{j,t})}{\Ls_{\mathrm{flow}}(H_{j,t})}\;
  \frac{\mathrm{d}}{\mathrm{d}\epsilon}\,\Ls_{\mathrm{flow}}\big(H_{j,t}+\epsilon\,\hat d\,\big)\Big|_{\epsilon=0}
\label{eq:utility}
\end{equation}
is the relative loss reduction per unit relative step, the hidden-state analogue of the gradient
angle of HASTE.

\textbf{Protocol.} $U$ is estimated in fp32 by symmetric central differences, with two partial
forward passes per cell and step size, so that the second-order term
$-\tfrac12\epsilon\,\mathrm{RMS}^2\,\hat d^\top\nabla^2\Ls\,\hat d/\Ls$ of a one-sided difference
does not enter; one-sided estimates turn whole regions with a weak first-order signal negative. The
readout $Q$ is fitted on the fit split, and $\hat d$ and $U$ are evaluated on a disjoint evaluation
split, so that readout overfitting cannot enter the intervention direction. Each cell uses at least
four noise realizations over several batches. Only cells whose bootstrap sign is stable, with a
95\% interval that excludes zero, are reported (59 of 60), and the conclusions hold for relative
step sizes $\epsilon\in\{0.02,0.05,0.10\}$.

\textbf{Controls.} (i)~The realizability control takes one alignment gradient step on the last
linear layer of block $j$, with every other parameter frozen, and measures the flow-loss change on
the same batch by central differences in parameter space; it agrees with $U$ in sign on 7 of 8
increment cells and on 8 of 8 whole-layer cells. (ii)~The function-class control retrains the
placement selected by $N\cdot[U]_+$, layer 3 at block 4, with the projector reduced from the
$3\times3$ convolution of iREPA to the $1\times1$ linear map of the readout, everything else
unchanged; its benefit rises from $+3.78$ to $+5.22$ gFID, still 16 gFID below the gap's choice.
The perturbation controls, (iii)~a norm-matched random direction and (iv)~a batch-shuffled teacher
target, both give $U\approx0$.

\appendixpart{II}{RARE}

\section{RARE Implementation Details}\label{app:impl}

The projector $P$ in Eq.~\ref{eq:objective} is the convolutional projector of iREPA, with
$3\times3$ kernels and 2048 hidden channels, and adds 0.9M parameters that are discarded after
training. The spatial normalization $\operatorname{sn}$ subtracts a fraction $\alpha=0.6$ of the
token mean and divides by the token standard deviation.

\appendixpart{III}{Experiments}

\section{Where Alignment Closes the Gap}\label{app:closure}

\begin{figure}[htbp]
\centering
\fpanel[0.4782\textwidth][0.125in]{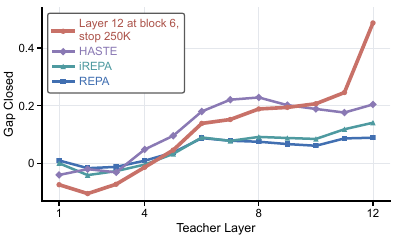}{a}%
\hfill
\fpanel[0.5216\textwidth][-0.060in]{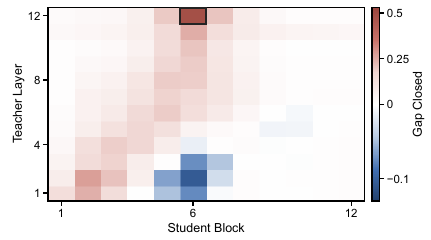}{b}%
\caption{\textbf{Alignment closes the gap where it is applied.} Gap closed relative to the
unaligned run at 200K steps. \textbf{(a)}~Along the teacher layers at block 6, for the run that
aligns layer 12 at block 6 and for REPA, iREPA and HASTE. \textbf{(b)}~Over the whole map for the
first run; the box marks the selected cell.}
\label{fig:closure_map}
\end{figure}

Section~\ref{sec:imagenet} reports that alignment closes the gap mainly at the layer it targets: by
0.49 at layer 12, against at most 0.27 elsewhere on the map. Figure~\ref{fig:closure_map} shows the
gap closed by the run that aligns the selected cell until 250K (Table~\ref{tab:ablation}, row 4),
relative to the unaligned run.

\section{Training and Selection Cost}\label{app:cost}

\begin{figure}[htbp]
\centering
\includegraphics[width=0.5181\textwidth]{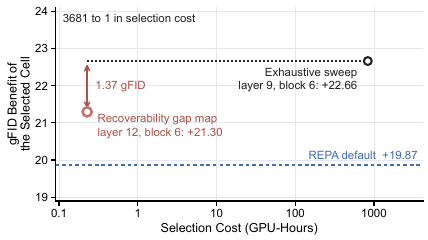}
\caption{\textbf{The map's cell comes within 1.37 gFID of the sweep's best at a fraction of its
cost.} Realized gFID benefit of the cell each procedure picks against the GPU-hours spent picking
it. Diagnosing once is $3681\times$ cheaper than the sweep on selection alone, and $19.9\times$
cheaper once the training of the selected cell is counted on both sides.}
\label{fig:cost}
\end{figure}

Table~\ref{tab:cost} details the training cost that Section~\ref{sec:efficiency} summarizes for the
runs of Table~\ref{tab:imagenet}, and Figure~\ref{fig:cost} compares the cell the map selects with
an exhaustive sweep that trains all twenty cells of the four-layer map to 100K steps.

\begin{table}[htbp]
\centering
\caption{\textbf{Training and inference cost} of the runs of Table~\ref{tab:imagenet}. Every method
discards its projector after training, so inference costs the same as for the unaligned model.
Marks rank the aligned methods only.}
\label{tab:cost}
\tablestyle
\setlength{\tabcolsep}{5pt}
\begin{tabular}{lrrrrr}
\toprule
 & \multicolumn{1}{c}{\hd{Unaligned}} & \multicolumn{4}{c}{\hd{Aligned}} \\
\cmidrule(lr){2-2}\cmidrule(lr){3-6}
 & \hd{SiT} & \hd{REPA} & \hd{iREPA} & \hd{HASTE} & \hd{RARE} \\
\midrule
Parameters (M)\down & 130.3 & 137.6 & \underline{135.6} & 137.6 & \textbf{131.2} \\
Peak Memory (GiB/GPU)\down & 6.22 & \underline{6.94} & \textbf{6.79} & 8.39 & 8.17 \\
Step Time (s)\down & 0.168 & 0.218 & 0.218 & \underline{0.212} & \textbf{0.188} \\
Training Cost (GPU-h to $400\mathrm{K}$)\down & 148.9 & 193.5 & 193.6 & \underline{188.0} & \textbf{167.1} \\
Inference Overhead & 0 & 0 & 0 & 0 & 0 \\
\bottomrule
\end{tabular}
\end{table}

\clearpage   
\section{Transfer: Full Results and Increment Targets}\label{app:transfer}

Table~\ref{tab:transfer_full} reports all metrics of the transfer study summarized in
Table~\ref{tab:transfer} (Section~\ref{sec:transfer}). Table~\ref{tab:transfer_inc} compares the
whole whitened target with the increment target in every setting; the increment arm aligns the cell
its own map selects, and the whole whitened layer has the lower FID everywhere.

\begin{table}[htbp]
\centering
\caption{\textbf{Transfer, all metrics.} (a,\,b) are unguided; $\Delta$FID is the improvement over
the unaligned SiT of the same setting. RARE aligns layer 9 on Places365. The better of iREPA and
RARE is in bold.}
\label{tab:transfer_full}
\tablestyle
\setlength{\tabcolsep}{4pt}
\begin{tabular}{llrrrrrrr}
\toprule
\hd{Setting} & \hd{Method} & \hd{FID\down} & \hd{sFID\down} & \hd{IS\up} & \hd{Prec.\up} & \hd{Rec.\up} & \hd{CMMD\down} & \hd{$\Delta$FID\up} \\
\midrule
\grouphead{9}{(a) Model Scale, DINOv2-B Teacher, 100K Steps}
SiT-B/2 & SiT & 62.59 & 7.15 & 20.78 & 0.395 & 0.567 & 1.703 & --- \\
 & iREPA & 37.70 & \textbf{6.95} & \textbf{40.19} & 0.514 & \textbf{0.618} & 1.408 & 24.89 \\
\rowcolor{oursrow}
 & \textbf{RARE} & \textbf{36.66} & 7.32 & 40.04 & \textbf{0.531} & 0.616 & \textbf{1.359} & \textbf{25.93} \\
\grouprule{9}
SiT-L/2 & SiT & 47.19 & 6.30 & 27.65 & 0.482 & 0.586 & 1.355 & --- \\
 & iREPA & 19.94 & 5.71 & 67.80 & 0.630 & 0.612 & 0.933 & 27.25 \\
\rowcolor{oursrow}
 & \textbf{RARE} & \textbf{18.93} & \textbf{5.58} & \textbf{68.17} & \textbf{0.644} & \textbf{0.624} & \textbf{0.881} & \textbf{28.26} \\
\grouprule{9}
SiT-XL/2 & SiT & 42.45 & 6.44 & 30.63 & 0.512 & 0.589 & 1.257 & --- \\
 & iREPA & 16.52 & 5.76 & \textbf{77.95} & 0.662 & 0.606 & 0.837 & 25.93 \\
\rowcolor{oursrow}
 & \textbf{RARE} & \textbf{16.07} & \textbf{5.55} & 76.79 & \textbf{0.663} & \textbf{0.617} & \textbf{0.804} & \textbf{26.38} \\
\midrule
\grouphead{9}{(b) Teacher, SiT-B/2, 100K Steps}
None & SiT & 62.59 & 7.15 & 20.78 & 0.395 & 0.567 & 1.703 & --- \\
\grouprule{9}
MAE-B/16 & iREPA & 49.95 & \textbf{6.70} & 26.93 & 0.463 & 0.586 & 1.496 & 12.64 \\
\rowcolor{oursrow}
 & \textbf{RARE} & \textbf{43.81} & 7.21 & \textbf{30.85} & \textbf{0.495} & \textbf{0.594} & \textbf{1.401} & \textbf{18.78} \\
\grouprule{9}
SigLIP-B/16 & iREPA & 40.60 & \textbf{7.02} & 36.50 & 0.496 & \textbf{0.619} & \textbf{1.445} & 21.99 \\
\rowcolor{oursrow}
 & \textbf{RARE} & \textbf{39.86} & 8.03 & \textbf{37.91} & \textbf{0.505} & 0.605 & 1.476 & \textbf{22.73} \\
\bottomrule
\end{tabular}

\vspace{7pt}
\setlength{\tabcolsep}{4pt}
\begin{tabular}{lrrrrcrrrr}
\toprule
\vhd{Method} & \multicolumn{4}{c}{\hd{Without Guidance}} & & \multicolumn{4}{c}{\hd{With Guidance}} \\
\cmidrule(lr){2-5}\cmidrule(lr){7-10}
 & \hd{FID\down} & \hd{KID$\times10^{3}$\down} & \hd{Prec.\up} & \hd{Rec.\up} & & \hd{FID\down} & \hd{KID$\times10^{3}$\down} & \hd{Prec.\up} & \hd{Rec.\up} \\
\midrule
\grouphead{10}{(c) Image Domain, Places365-Standard, SiT-B/2, 400K Steps}
SiT & 11.40 & 7.89{\scriptsize$\,\pm0.91$} & 0.565 & 0.535 & & 6.79 & 3.06{\scriptsize$\,\pm0.46$} & 0.626 & 0.524 \\
\grouprule{10}
iREPA & 7.71 & 4.50{\scriptsize$\,\pm0.64$} & \textbf{0.608} & \textbf{0.565} & & 4.98 & 1.83{\scriptsize$\,\pm0.31$} & 0.652 & \textbf{0.556} \\
\rowcolor{oursrow}
\textbf{RARE} & \textbf{7.33} & \textbf{4.38}{\scriptsize$\,\pm0.63$} & 0.606 & 0.563 & & \textbf{4.54} & \textbf{1.51}{\scriptsize$\,\pm0.27$} & \textbf{0.666} & 0.547 \\
\bottomrule
\end{tabular}
\end{table}

\begin{table}[htbp]
\centering
\caption{\textbf{Whitened layer against increment target.} FID-50K without guidance, 100K steps on
ImageNet and 400K on Places365. Each increment arm uses the cell its own map selects, with the same
projector and loss weight. The better target is in bold.}
\label{tab:transfer_inc}
\tablestyle
\setlength{\tabcolsep}{4pt}
\begin{tabular}{llcrlcr}
\toprule
\vhd{Setting} & \multicolumn{3}{c}{\hd{Whitened Layer}} & \multicolumn{3}{c}{\hd{Increment}} \\
\cmidrule(lr){2-4}\cmidrule(lr){5-7}
 & \hd{Layer} & \hd{Block} & \hd{FID\down} & \hd{Layer} & \hd{Block} & \hd{FID\down} \\
\midrule
SiT-B/2, ImageNet & Layer 12 & 4 & \textbf{36.66} & Layer 12 & 6 & 39.21 \\
SiT-L/2, ImageNet & Layer 12 & 8 & \textbf{18.93} & Layer 9 & 4 & 24.26 \\
SiT-XL/2, ImageNet & Layer 12 & 8 & \textbf{16.07} & Layer 9 & 5 & 20.15 \\
\grouprule{7}
SiT-B/2, Places365 & Layer 9 & 4 & \textbf{7.33} & Layer 9 & 2 & 9.19 \\
\bottomrule
\end{tabular}
\end{table}

\section{Other Backbones}\label{app:backbones}

The two backbones of Section~\ref{sec:transfer} each change one part of the setup: DiT the
generative objective and its sampler, MM-DiT the conditioning architecture.

\textbf{DiT.} DiT \citep{dit} replaces flow matching with $\epsilon$-prediction and an ADM
linear-beta DDPM sampler. Every aligned method keeps its published target and depth; only the
generative objective and the sampler change. All aligned methods improve on the unaligned DiT, in
the same order as in Table~\ref{tab:imagenet}, and RARE has the lowest FID with and without guidance
(Table~\ref{tab:dit}).

\textbf{MM-DiT.} MM-DiT keeps one set of weights per modality and joins them with a single
attention softmax \citep{sd3}. We instantiate it at the shape of DiT-B/2, with twelve blocks, 768
channels, twelve heads and patch size 2; the class embedding enters twice, pooled into the adaLN
vector and expanded into four learned context tokens that share attention with the 256 image
tokens. Duplicating the transformer per modality makes the backbone larger than DiT-B/2, 250.4M
against 130.5M parameters, so we do not compare absolute FID across tables. The generative
objective stays the flow matching of Table~\ref{tab:imagenet}, so this comparison isolates the
conditioning architecture, and alignment reads only the image tokens (Table~\ref{tab:mmdit}).

\begin{table}[htbp]
\centering
\caption{\textbf{DiT-B/2 with $\epsilon$-prediction and a 250-step DDPM sampler, 400K steps.}
Notation as in Table~\ref{tab:imagenet}.}
\label{tab:dit}
\tablestyle
\setlength{\tabcolsep}{4pt}
\begin{tabular}{lllrrrrrr}
\toprule
\vhd{Method} & \multicolumn{2}{c}{\hd{Alignment}} & \vhd{FID\down} & \vhd{sFID\down} & \vhd{IS\up} & \vhd{Prec.\up} & \vhd{Rec.\up} & \vhd{CMMD\down} \\
\cmidrule(lr){2-3}
 & \hd{Target} & \hd{Block} & & & & & & \\
\midrule
\grouphead{9}{Without Guidance}
\rowcolor{baserow}
DiT & --- & --- & 40.50 & \underline{6.26} & 35.05 & 0.502 & 0.631 & 1.341 \\
\grouprule{9}
REPA & L12 & 4 & 30.05 & 6.35 & 49.86 & 0.556 & \underline{0.650} & 1.203 \\
iREPA & L12 & 4 & 24.16 & 6.36 & 63.68 & 0.586 & 0.643 & --- \\
HASTE & L12 + attn. & 8 & \underline{21.76} & \underline{6.26} & \underline{65.55} & \underline{0.612} & 0.629 & \underline{1.030} \\
\grouprule{9}
sREPA & L12 + Gram & 4 & 27.61 & 6.27 & 54.90 & 0.568 & \textbf{0.651} & 1.167 \\
\midrule
\rowcolor{oursrow}
\textbf{RARE (Ours)} & L12, whitened & 4 & \textbf{20.54} & \textbf{6.00} & \textbf{68.14} & \textbf{0.619} & 0.631 & \textbf{0.995} \\
\midrule
\grouphead{9}{With Guidance (scale 1.65 on [0,\,0.72])}
\rowcolor{baserow}
DiT & --- & --- & 13.23 & \underline{5.04} & 101.05 & 0.736 & 0.495 & 0.910 \\
\grouprule{9}
REPA & L12 & 4 & 7.45 & 5.14 & 153.65 & 0.773 & \underline{0.505} & 0.817 \\
iREPA & L12 & 4 & 5.06 & 5.24 & 195.60 & 0.806 & 0.500 & 0.769 \\
HASTE & L12 + attn. & 8 & \underline{4.90} & 5.16 & \underline{197.91} & \underline{0.833} & 0.470 & \underline{0.680} \\
\grouprule{9}
sREPA & L12 + Gram & 4 & 6.22 & 5.19 & 170.43 & 0.790 & \textbf{0.506} & 0.785 \\
\midrule
\rowcolor{oursrow}
\textbf{RARE (Ours)} & L12, whitened & 4 & \textbf{4.67} & \textbf{5.02} & \textbf{205.32} & \textbf{0.847} & 0.465 & \textbf{0.641} \\
\bottomrule
\end{tabular}
\end{table}

\begin{table}[htbp]
\centering
\caption{\textbf{MM-DiT-B/2 with flow matching, 100K steps.} Notation as in
Table~\ref{tab:imagenet}. A tie for best leaves no second best.}
\label{tab:mmdit}
\tablestyle
\setlength{\tabcolsep}{4pt}
\begin{tabular}{lllrrrrrr}
\toprule
\vhd{Method} & \multicolumn{2}{c}{\hd{Alignment}} & \vhd{FID\down} & \vhd{sFID\down} & \vhd{IS\up} & \vhd{Prec.\up} & \vhd{Rec.\up} & \vhd{CMMD\down} \\
\cmidrule(lr){2-3}
 & \hd{Target} & \hd{Block} & & & & & & \\
\midrule
\grouphead{9}{Without Guidance}
\rowcolor{baserow}
MM-DiT & --- & --- & 56.19 & 7.45 & 25.88 & 0.419 & 0.593 & 1.688 \\
\grouprule{9}
REPA & L12 & 4 & 42.79 & \underline{7.04} & 35.11 & 0.481 & 0.620 & 1.467 \\
iREPA & L12 & 4 & 35.34 & \textbf{6.79} & 44.17 & 0.518 & \textbf{0.630} & 1.378 \\
HASTE & L12 + attn. & 8 & \underline{34.28} & 7.38 & \underline{45.10} & \textbf{0.537} & \underline{0.628} & \underline{1.332} \\
\midrule
\rowcolor{oursrow}
\textbf{RARE (Ours)} & L12, whitened & 4 & \textbf{32.98} & 7.07 & \textbf{47.48} & \textbf{0.537} & 0.615 & \textbf{1.313} \\
\midrule
\grouphead{9}{With Guidance (scale 1.65 on [0,\,0.72])}
\rowcolor{baserow}
MM-DiT & --- & --- & 33.34 & 6.27 & 48.45 & 0.534 & 0.559 & 1.440 \\
\grouprule{9}
REPA & L12 & 4 & 22.48 & \underline{5.94} & 71.79 & 0.605 & \underline{0.583} & 1.234 \\
iREPA & L12 & 4 & \underline{15.59} & \textbf{5.72} & \underline{97.31} & \underline{0.650} & 0.577 & 1.136 \\
HASTE & L12 + attn. & 8 & 16.54 & 6.31 & 96.38 & 0.644 & \textbf{0.588} & \underline{1.127} \\
\midrule
\rowcolor{oursrow}
\textbf{RARE (Ours)} & L12, whitened & 4 & \textbf{13.76} & 6.01 & \textbf{105.83} & \textbf{0.679} & 0.562 & \textbf{1.071} \\
\bottomrule
\end{tabular}
\end{table}

\section{Qualitative Results}\label{app:qualitative}

\begin{figure}[htbp]
\centering
\includegraphics[width=\textwidth]{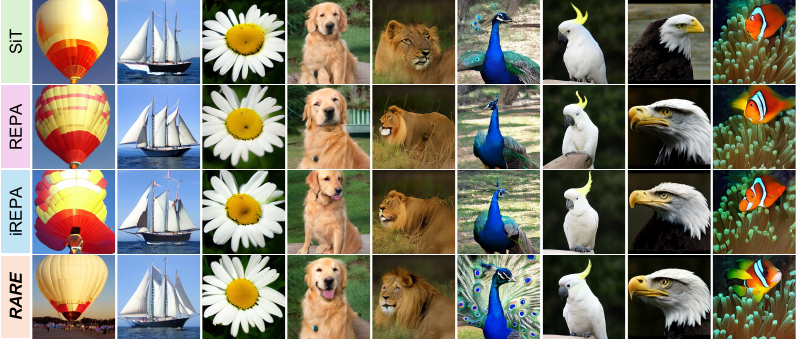}
\caption{\textbf{ImageNet-256 samples at 400K steps.} Each column shares one class, initial noise
and sampler noise across the rows, so differences within a column come from the model; guidance
4.0.}
\label{fig:qualitative}
\end{figure}

\begin{figure}[htbp]
\centering
\includegraphics[width=\textwidth]{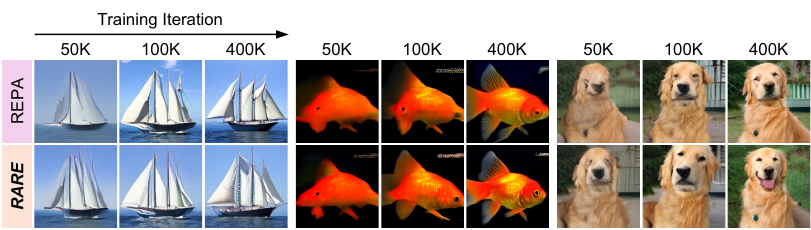}
\caption{\textbf{REPA and RARE over training.} Three classes at 50K, 100K and 400K steps; guidance 4.0.}
\label{fig:iteration}
\end{figure}

\begin{figure}[htbp]
\centering
\includegraphics[width=0.6\textwidth]{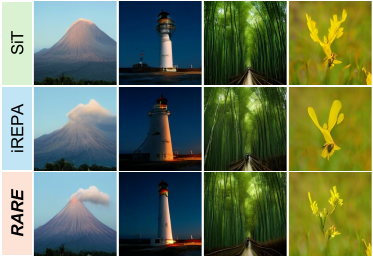}
\caption{\textbf{Places365-Standard at 400K steps.} SiT, iREPA and RARE on shared columns; guidance 4.0.}
\label{fig:places}
\end{figure}

\begin{figure}[htbp]
\centering
\includegraphics[width=0.84\textwidth]{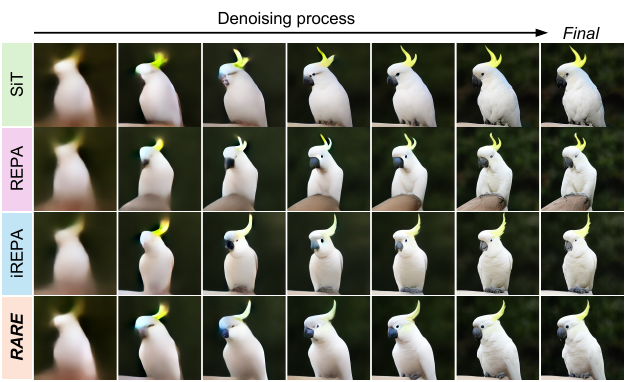}
\caption{\textbf{One column through the sampler.} Decoded prediction of the clean latent at seven
points of the 250-step SDE; guidance 4.0. The rows share the noise and the first snapshot and separate at the second, where the three aligned
rows already resolve the eye and the beak.}
\label{fig:trajectory}
\end{figure}

The samples below accompany Sections~\ref{sec:imagenet} and~\ref{sec:transfer}.
Figure~\ref{fig:qualitative} compares the four methods at 400K steps, Figure~\ref{fig:iteration}
follows REPA and RARE over training, Figure~\ref{fig:places} shows Places365 samples, and
Figure~\ref{fig:trajectory} follows one column through the sampler.

\end{document}